\documentclass[letterpaper, 10 pt, conference]{ieeeconf}  

\IEEEoverridecommandlockouts                              

\makeatletter

\newcommand{\Rmnum}[1]{\expandafter\@slowromancap\romannumeral #1@}
\makeatother
\makeatletter
\let\NAT@parse\undefined
\makeatother
\usepackage[pagebackref=false,breaklinks=true,colorlinks=true,bookmarks=false,linkcolor={red!50!black},urlcolor={blue},citecolor={green!60!black}]{hyperref} 

\usepackage{cite}
\usepackage{url}
\usepackage{pifont}
\usepackage{booktabs}
\usepackage{etoolbox}
\usepackage{float}
\usepackage{comment}

\makeatletter
\patchcmd{\@makecaption}
  {\scshape}
  {}
  {}
  {}
\makeatletter
\patchcmd{\@makecaption}
  {\\}
  {.\ }
  {}
  {}
\makeatother
\usepackage{pifont}
\usepackage{rotating}
\usepackage{amssymb}
\usepackage{amsmath}
\usepackage{CJKutf8}
\usepackage{booktabs}
\usepackage{multirow} 
\usepackage{mathrsfs}
\usepackage{graphicx}
\usepackage{etoolbox}
\usepackage{caption}
\usepackage{color}
\usepackage{bm}
\usepackage[table,dvipsnames]{xcolor}
\usepackage{bbding}

\definecolor{gray}{rgb}{0.92, 0.92, 0.92}
\definecolor{yellow}{rgb}{1, 1, 0.7}
\definecolor{orange}{rgb}{1, 0.85, 0.7}
\definecolor{red}{rgb}{1, 0.7, 0.7}

\title{\LARGE \bf
DGTR: Distributed Gaussian Turbo-Reconstruction for\\ Sparse-View Vast Scenes
}

\author{Hao Li $^{1,2}$, Yuanyuan Gao$^{1}$, Haosong Peng$^{3}$,  Chenming Wu$^2$, Weicai Ye$^4$,  Yufeng Zhan$^{3}$, \\ Chen Zhao$^2$, Dingwen Zhang$^{1, \dagger}$, Jingdong Wang$^2$~\IEEEmembership{Fellow, IEEE}, Junwei Han$^1$~\IEEEmembership{Fellow, IEEE} 
}

\begin{document}

 \twocolumn[{
 \renewcommand\twocolumn[1][]{#1}
\maketitle
 \thispagestyle{empty}
 \begin{center}
     \captionsetup{type=figure}
     \vspace{-10pt} \includegraphics[width=0.91\linewidth]{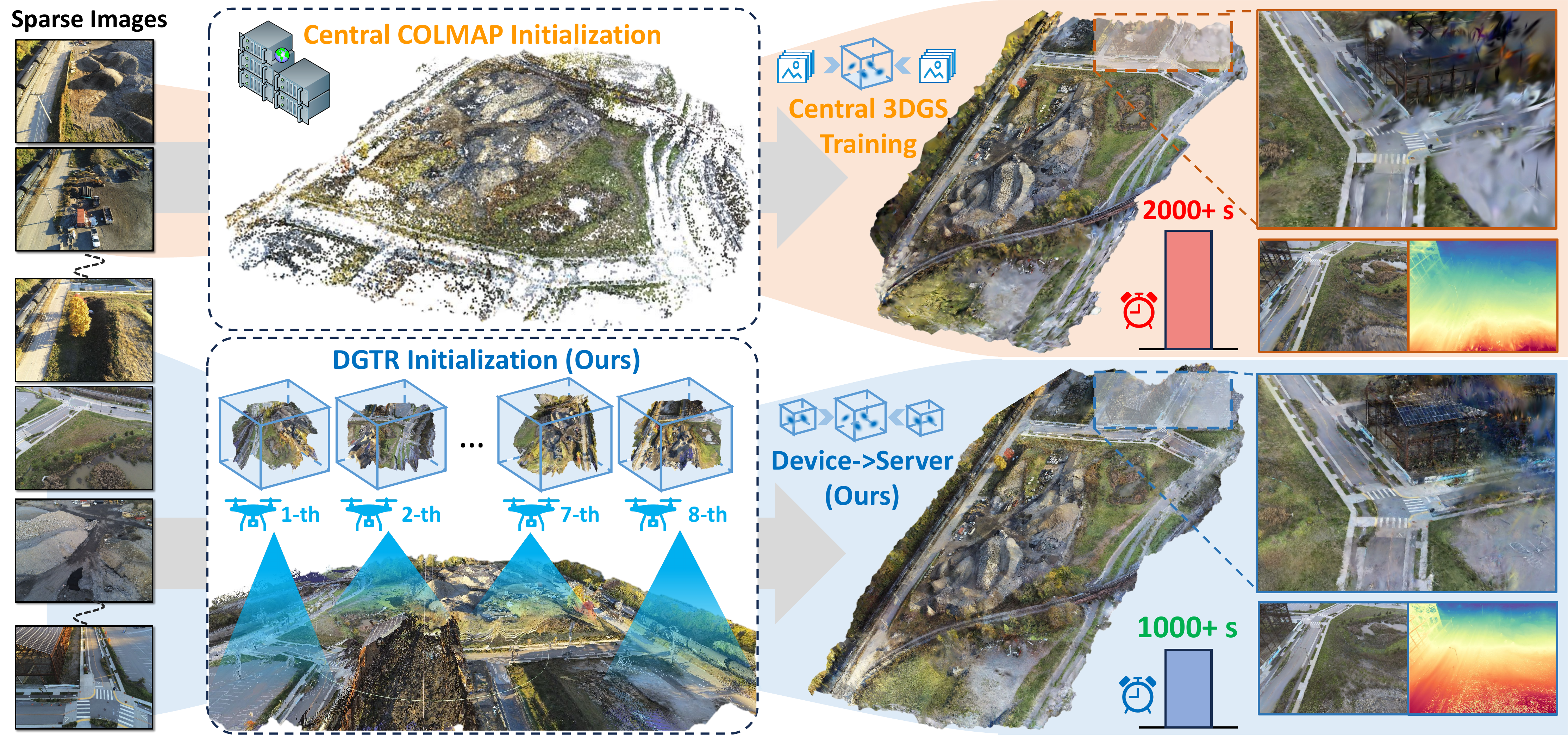}
     \captionof{figure}{Our proposed \textbf{DGTR} can rapidly reconstruct sparse-view vast scenes in a distributed manner. Compared with the standard central 3DGS training method, we achieve better visual appearance and geometry accuracy at a faster speed.}
     \label{fig:teaser}
 \end{center}
 }]
{
  \footnotetext[1]{BRAIN Lab, NWPU, China. {\tt\small lifugan\_10027@outlook.com  \{li0jingfeng, zhangdingwen2006yyy\}@gmail.com}}
  \footnotetext[2]{Baidu VIS, Beijing, China. {\tt\small \{wuchenming, zhaochen02, wangjingdong\}@baidu.com}}
 \footnotetext[3]{X-NS Group, Beijing Institute of Technology, Beijing, China. {\tt\small \{livion, yu-feng.zhan\}@bit.edu.cn}}
 \footnotetext[4]{State Key Lab of CAD \& CG, Zhejiang University, Hangzhou, China. {\tt\small maikeyeweicai@gmail.com }
  \renewcommand{\thefootnote}{\fnsymbol{footnote}}
 \footnotetext[2]{Corresponding author. H. Li, Y. Gao, and H. Peng contributed equally.}
}

\begin{abstract}


Novel-view synthesis (NVS) approaches play a critical role in vast scene reconstruction. However, these methods rely heavily on dense image inputs and prolonged training times, making them unsuitable where computational resources are limited. Additionally, few-shot methods often struggle with poor reconstruction quality in vast environments.
This paper presents DGTR, a novel distributed framework for efficient Gaussian reconstruction for sparse-view vast scenes. Our approach divides the scene into regions, processed independently by drones with sparse image inputs. Using a feed-forward Gaussian model, we predict high-quality Gaussian primitives, followed by a global alignment algorithm to ensure geometric consistency. Synthetic views and depth priors are incorporated to further enhance training, while a distillation-based model aggregation mechanism enables efficient reconstruction.
Our method achieves high-quality large-scale scene reconstruction and novel-view synthesis in significantly reduced training times, outperforming existing approaches in both speed and scalability. We demonstrate the effectiveness of our framework on vast aerial scenes, achieving high-quality results within minutes. Code will released on our project page \href{https://3d-aigc.github.io/DGTR}{https://3d-aigc.github.io/DGTR}.
\end{abstract}

\section{INTRODUCTION}

Differentiable volume rendering techniques, such as Neural Radiance Field (NeRF)~\cite{mildenhall2021nerf} and 3D Gaussian Splatting (3DGS)~\cite{kerbl20233d}, have become instrumental in advancing 3D vision and robotics. These methods are crucial for a wide range of applications that require synthesizing novel views, including autonomous driving~\cite{waymo_open_dataset}, robotic simulation~\cite{byravan2023nerf2real}, and embodied intelligence~\cite{kwon2023renderable}. By enabling novel view synthesis (NVS), they allow robots to ``imagine" scenes from different perspectives, much like human vision does.

A typical approach for novel view synthesis~\cite{li2024dngaussian,zhu2023fsgs} involves using dense image inputs to cover as much scene area as possible, along with extended training times to achieve high-quality results. However, in large-scale scene synthesis scenarios, particularly with aerial imagery, drones and UAVs face significant limitations in terms of computational power and storage capacity. These constraints make it challenging to collect and process dense image datasets required for training~\cite{gao2024multiplane}.
In contrast, another line of approaches uses sparse views (i.e., few-shot) as input. These methods are designed to synthesize vast scenes, such as those used in aerial photogrammetry, from limited image data. However, existing large-scale scene reconstruction techniques~\cite{lin2024vastgaussian,chen2024dogaussian} primarily focus on addressing memory constraints and expanding reconstruction areas. They struggle to handle few-shot inputs, leading to poor reconstruction quality in localized regions with a limited number of available images.

Existing methods, such as VastGaussian~\cite{lin2024vastgaussian} and DoGaussian~\cite{chen2024dogaussian}, describe themselves as distributed approaches. However, they still rely on a global COLMAP initialization and require the division of global point clouds and images across devices. This means they are not suitable for truly distributed systems, where devices need to perform reconstruction tasks independently. At the same time, current few-shot methods~\cite{wang2023sparsenerf, yang2023freenerf, li2024dngaussian, zhu2023fsgs} are limited to small scenes or object-level tasks, typically using only 2 or 3 images as input. These methods fall short when applied to the complexities of large-scale aerial scenes, leaving a gap with the goal of this work. Additionally, methods like FSGS~\cite{zhu2023fsgs} and DNGaussian~\cite{li2024dngaussian} use depth regularization as an additional form of supervision. Unfortunately, despite this, they still struggle to achieve high-quality novel view synthesis due to inadequate Gaussian initialization.

To overcome the aforementioned issues, we establish a novel distributed 3D Gaussian rendering framework for few-shot vast scene reconstruction. 
With several drones, each drone is responsible for a non-overlap region, capturing sparse images, initializing Gaussian points, and training the local Gaussian model. This strategy enables drones can perform Gaussian training and NVS tasks in their own regions. With fewer images, the initialization and training time are greatly reduced instead of waiting for global COLMAP and points division.
Meanwhile, we propose a novel initialization strategy by leveraging the pre-trained feed-forward Gaussian method and a global alignment algorithm to align the positions and scales for all generated Gaussian points. It dramatically boosts both the quality and the convergence time with better geometry and Gaussian attributes. Subsequently, we employ depth regularization supervision to prevent overfitting of the Gaussians and enhance their accuracy from the novel view.
Finally, we introduce a distillation-based model aggregation algorithm to obtain the final high-quality large-scale scene model. Instead of transmitting captured images, each drone uploads the Gaussian model to the central server. Throughout the aggregation process, the local models synthesize training view images as pseudo GT images, and then the central server merges the local models as a student model and is supervised by the pseudo-GT images.
Our main contributions are summarized as follows:
\begin{enumerate}
    \item We propose a novel distributed framework for sparse-view vast scene reconstruction, enabling each drone to independently perform reconstruction and NVS tasks, significantly improving both speed and quality.
    \item We deploy a feed-forward Gaussian method with global alignment and depth regularization, ensuring accurate geometry and fast convergence with few-shot supervision.
    \item We present the first sparse-view vast scene dataset and benchmark, to the best of our knowledge,  using key-frame sampling from large-scale datasets.
\end{enumerate}
Our method outperforms both distributed and centralized approaches, enabling large-scale scene reconstruction in minutes, making it practical for real-world use.


\section{Related Work}

\subsection{Few-shot Novel View Synthesis}
Few-shot novel view synthesis is particularly important yet challenging due to the scarcity of data and the need to generate photo-realistic results for unseen views of various applications such as AR / VR, autonomous driving, and embodied AI.
Traditional few-shot novel view synthesis methods~\cite{du2023learning,niemeyer2022regnerf,yang2023freenerf,li2024dngaussian,zhu2023fsgs} can only generate novel views on small-scale scenes~\cite{mildenhall2019local} or object-level datasets~\cite{deitke2023objaverse}. 
Latest methods, such as DNGaussian~\cite{li2024dngaussian} and FSGS~\cite{zhu2023fsgs}, address the sparse-view problem by introducing regularization strategies. Furthermore, the newly proposed MvSplat~\cite{chen2024mvsplat} and ReconX~\cite{liu2024reconx} incorporate the generative model and the diffusion process respectively to overcome the information loss from the sparse-view sampling condition.
Nevertheless, the problem of sparse-view synthesis in the field of large-scale scenes remains uncharted, where the onboard computation resource on the edge observer, say a drone, is usually insufficient to conduct heavy calculations, and it is difficult to keep high-quality image communication to the central server. 
%

\subsection{Feed-Forward 3D Gaussian Splatting}

Apart from per-scene optimization methods, a growing number of feed-forward 3DGS models~\cite{charatan2024pixelsplat,chen2024mvsplat,szymanowicz2024flash3d,szymanowicz2024splatter,zheng2024gps,smart2024splatt3r, li2024ggrt} have been proposed to directly generate 3D Gaussian maps using Transformer~\cite{vaswani2017attention} networks from few views without training Gaussian attributes.
These methods leverage data-driven priors acquired from training vision foundation models on large datasets of scenes, enabling them to capture Gaussian primitives from sparse views more effectively.
%
%
For example, by MvSplat~\cite{chen2024mvsplat}, DUSt3R ~\cite{wang2024dust3r} achieves significant performance improvement on stereo point cloud estimation using uncalibrated images.
Building upon DUSt3R, Splatt3R~\cite{smart2024splatt3r} further predicts the additional Gaussian attributes for each point by incorporating an additional novel view synthesis loss in the pre-train procedure, thus achieving high visual quality and perceptual similarity to the ground truth images.
However, such methods are unable to reconstruct vast scenes as the fact that they can only accept at most 5 views~\cite{chen2024mvsplat} as input and difficult to perform global optimization across all views.
%

\begin{figure*}[t]
\begin{center}
\includegraphics[width=\linewidth]{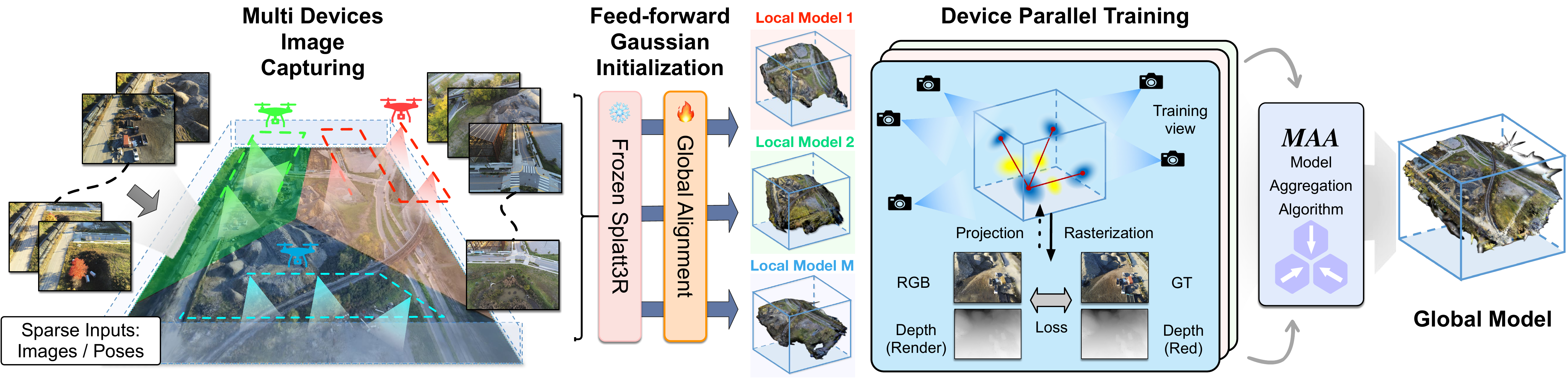}
\caption{\textbf{DGTR (Ours) Overview:} Given \(M\) individual devices (drones), we aim to perform sparse-view vast scene reconstruction with fast speed in a multi-device collaboration manner.  The whole pipeline can be divided into three steps: 1) each device explores a non-overlap region and conducts Gaussian initialization using the off-the-shelf feed-forward Gaussian method and global alignment strategy; 2) each device performs sparse-view scene reconstruction using the initialized Gaussians; 3) The device uploads the well-trained Gaussian model to the central server, the central server performs model aggregation in a distillation manner. }
\label{fig:overview}
\end{center}
\vspace{-8mm}
\end{figure*}

\subsection{Large Scale Reconstruction}
%
Traditional methods~\cite{agarwal2011building, fruh2004automated, schonberger2016structure} employed a structure-from-motion (SfM) framework to ascertain camera positions and produce sparse point clouds. 
%
%
More recently, advancements such as NeRF~\cite{mildenhall2021nerf} and 3DGS~\cite{kerbl20233d} have emerged as a worldwide 3D representation system thanks to their photo-realistic characteristics and the ability of novel-view synthesis, which inspires many works~\cite{zhenxing2022switch, zhang2023efficient, turki2022mega, tancik2022block, lin2024vastgaussian, xiangli2022bungeenerf, xu2023grid, li2024dgnr, liu2024citygaussian, li2024ho, chen2024dogaussian,suzuki2024fed3dgs} to extend it into vast scene reconstructions.
As the scale of reconstruction scenes continually expands, the limited video memory of the server and the prohibitive computational overhead restrict their capacity for rapid reconstruction. 
To address this, the above methods can be categorized into centralized~\cite{zhenxing2022switch, turki2022mega, tancik2022block, lin2024vastgaussian,  liu2024citygaussian, chen2024dogaussian} and distributed~\cite{suzuki2024fed3dgs} framework.
Most centralized approaches~\cite{lin2024vastgaussian,liu2024citygaussian,chen2024dogaussian} based on 3DGS adopt a divide-and-conquer strategy, gathering all images onto a single server and implement a data (or region) partitioning strategies to distribute the workload to multiple GPUs.
For the distributed approach, recently proposed Fed-3DGS~\cite{suzuki2024fed3dgs} distributed reconstruction tasks across numerous edge devices, which independently collect images and the conduct the training pipeline. 
However, both centralized and distributed frameworks suffer from dense camera views and slow convergence rates, which often require several hours or even days of reconstruction training.
In contrast, our proposed DGTR is the first to employ a well-designed algorithm upon the distributed framework, realizing rapid reconstruction in minutes.

\section{Preliminaries}
Gaussian Splatting represents a set of 3D Gaussian primitives $\mathbb{G} = \{\mathbf{G}_i\}$.
Each 3D Gaussian $\mathbf{G}_i = (\mathbf{x}_i, \mathbf{\Sigma}_i, \mathbf{S}_i, \alpha_i)$ is characterized by its position $\mathbf{x}_i=(x_i, y_i, z_i)$, covariance matrix $\mathbf{\Sigma}_i\in\mathbb{R}^{3\times3}$, opacity $\alpha_i\in\mathbb{R}$, and spherical harmonic coefficients $\mathbf{S}_i\in\mathbb{R}^{3\times d}$ for view-dependent colors, where $d$ is related to the degree of the spherical harmonics. Notably, covariance matrix $\mathbf{\Sigma}_i$ can be decomposed into the rotation matrix $\bm{R}_k\in \mathbb R^{3\times3}$ and the scaling matrix $\bm{S}_k\in \mathbb R^{3\times3}$ with the following equation:
\begin{equation}
\mathbf{\Sigma}_i = \mathbf{R}_i\mathbf{S}_i\mathbf{S}_i^\top \mathbf{R}_i^\top.
\end{equation}
These primitives parameterize the 3D radiance field of the scene, through a tiled-based rasterization process, 3DGS facilitates real-time alpha blending of numerous Gaussians to render novel-view images.

\section{Methodology}
Our goal is to reconstruct the entire vast scene in a distributed 3DGS manner quickly. 
Given $\mathbf{M}$ devices, each device \(m\) explores in the non-overlapping region and captures several sparse images  $\mathcal{I}_m = \{\mathbf{I}^i\}_{i=\{1,\ldots,N_m\}}$ $(\mathbf{I}^i\in\mathbb{R}^{H\times W\times 3})$ and corresponding poses $\mathcal{P}_m = \{\mathbf{P}^i\}_{i=\{1,\ldots,N_m\}}$ $(\mathbf{P}^i\in\mathbb{R}^{4\times 4})$.
Then we perform Gaussian \(\mathbb{G}_m\) initialization using the feed-forward Gaussian model instead of COLMAP (see Sec.~\ref{sec:gi}) followed by Sparse-view Gaussian Training Procedure (Sec.~\ref{sec:dpt}) to get the final local Gaussian model \(\hat{\mathbb{G}}_m\).
Upon completion of training, each device uploads its local model \(\hat{\mathbb{G}}_m\) to the central server. 
Once the central server receives all local models from the devices, it aggregates them into a unified global model \(\hat{\mathbb{G}}_g\). We outline our model aggregation method in Sec.~\ref{sec:fa}.

\subsection{Gaussian Initialization}\label{sec:gi}
Gaussian initialization is a fundamental step in the training pipeline. 
A high-quality Gaussian initialization will enhance training efficiency and facilitate rapid convergence.
\subsubsection{Feed-forward Gaussian Model Inference}\label{sec:vfmi}
The original 3DGS uses SfM to initialize the Gaussian primitives of the scene, which takes enormous optimized time to produce sparse points and is unable to initialize all primitives of the Gaussian points.
Nowadays, feed-forward 3DGS~\cite{smart2024splatt3r,chen2024mvsplat,li2024ggrt} have been proven to be a reliable way to reconstruct scenes in sparse input views.
Inspired by this, we propose to use the pre-trained feed-forward Gaussian model (\textit{i.e.} Splatt3R~\cite{smart2024splatt3r}) as the predictor for each device, which takes the collected images \(\mathcal{I}_m\) as input and outputs the high-quality Gaussian primitives \(\mathbb{G}_m\).
Hence Splatt3R uses image pairs as input, here we perform sliding windows to obtain image pairs $(\{\mathbf{I}^i_m, \mathbf{I}^{i+1}_m\}, i=1,\ldots,N_m-1)$ with co-visible areas from images \(\mathcal{I}_m\).
Specifically, for \(i\)-th image pair $\{\mathbf{I}^i_m, \mathbf{I}^{i+1}_m\}$, the feed-forward Gaussian model predicts pixel-aligned 3D Gaussian primitives \(\mathbb{G}_m^i\) as shown:
\begin{equation}
\begin{aligned}
& \mathbb{G}_m^i=\text { Feed-forward }\left(\mathbf{I}_m^i, \mathbf{I}_m^{i+1}\right) \\
& \mathbb{G}_m^i=\left\{\left(c_k, \mathbf{x}_k, \mathbf{\Sigma}_k, \mathbf{S}_k, \alpha_k\right)_{k=1}^K\right\},
\end{aligned}
\end{equation}
where \(c\) is the confidence score, \(K=2\times H \times W\) and \(H,W\) denote the height and weight of the input images.

\subsubsection{Global Alignment}\label{sec:ga}
Due to Splatt3R's pose-free nature, the Gaussian points it predicts are scale-irrelevant. 
Therefore, we must resolve scale ambiguity from estimated to GT poses to recover a global-aligned Gaussian map. 
To achieve this, we introduce an optimized-based global alignment strategy for both the scale of position \(\mathbf{x}\) and the scale of covariance \(\mathbf{\Sigma}\) in Gaussian attributes, building upon the foundation of  DUSt3R~\cite{wang2024dust3r}.

Specifically, we first construct a connectivity graph $\mathcal{G} (\mathcal{V}, \mathcal{E})$ from images \(\mathcal{I}_m\), meanwhile $N_m$ images form vertices $\mathcal{V}$ and each edge $e = (p, q) \in \mathcal{E}$ indicates that images $\mathbf{I}^p$ and $\mathbf{I}^q$ has co-visible areas. We then extract \(c_i\) and \(x_i\) from Gaussian primitives predicted by Splatt3R and form confidence maps \(C^{p,p}, C^{q,p}\) and pointmaps \(X^{p,p}, X^{q,p}\).
To align the global pointmaps $\{\mathcal{X}^n\in\mathbb{R}^{W\times H\times3}\}$ for all images \(n=1,\cdots,N\), we then formulate the following optimization problem to minimize the 3D-projection error by introducing parameterized pointmaps \(\mathcal{X}\) and scaling $\sigma_e> 0$ associated to each pair $e$:
\begin{equation}
    \chi^*=\underset{\chi,\sigma}{\text{argmin}}\sum_{e\in\mathcal{E}}\sum_{v\in e}\sum_{i=1}^{HW}C_i^{v,e}\left\|\chi_i^v-\sigma_eP_eX_i^{v,e}\right\|,
\end{equation}
where pose $P_e \in \mathbb{R}^{3×4}$ are given by the drone.  After that, we replace the position \(\mathbf{x}\)  with optimized pointmaps \(\mathcal{X}\) for all Gaussian points and form the refined Gaussian set \(\hat{\mathbb{G}}_m^i\).

For the scale of covariance \(\mathbf{\Sigma}\), we simply adopt the classical ICP (Iterative Closest Point) algorithm to predict the global scale \(s^g_m\in \mathbb{R}^{1}\) between original Gaussians  \(\mathbb{G}_m^i\) to optimized Gaussians  \(\hat{\mathbb{G}}_m^i\). Nevertheless, ICP can only update the global scale, ignoring the relative scale \(s^r_m\in \mathbb{R}^{2\times H\times W}\)  between the image-level pointmaps, leading to insufficient Gaussian representation for both near and far regions. Specifically, we rescale the scale by measuring the average distance between adjacent points to ensure good coverage of the 3D scene and eliminate Gaussian with abnormal scale. The final Gaussian set \(\widetilde{\mathbb{G}}_m^i\) can be expressed below:
\begin{align}
    \widetilde{\mathbb{G}}_m^i = s^g_m \times s^l_m \times \hat{\mathbb{G}}_m^i.
\end{align}

After the above steps, we concatenate all the pairs' Gaussian sets  \( \widetilde{\mathbb{G}}_m^i \),  obtaining fine-grained 3D Gaussian initialization \(\mathbb{G}_m\) for region reconstruction.

\subsection{Multiple Devices Parallel Training}\label{sec:dpt}
The device begins to train the model using a partial image set.
We use additional depth information from the training views to address the inherent issue of overfitting to sparse training views to supervise the model training.
Here a relaxed relative Pearson correlation is used as loss \(\mathcal{L}_{reg}\) to measures the distribution difference between 2D depth maps:
\begin{equation}
    \mathcal{L}_{reg}(\mathbf{D}, \mathbf{D}_{\mathrm{est}})=\left\|\frac{\mathrm{Cov}(\mathbf{D}, \mathbf{D}_{\mathrm{est}})}{\sqrt{\mathrm{Var}(\mathbf{D})\text{Var}(\mathbf{D}_{\mathrm{est}})}}\right\|_1,
\end{equation}
where \(\mathbf{D}\) is the depth map rendered by Gaussian model,  $\mathbf{D}_{est}$ is the depth maps predicted by pre-trained depth estimator DPT~\cite{ranftl2021vision}, \(\mathrm{Cov}(\cdot, \cdot)\) and \(\mathrm{Var}(\cdot, \cdot)\) denote covariance and variance separately.
This soften constraint allows for the alignment of depth structure without being hindered by the inconsistencies in absolute depth values.

To enable the backpropagation from depth before guiding Gaussian training, we implement a differentiable depth rasterizer, allowing for receiving the error signal between the rendered depth $D_{ras}$ and the estimated depth $D_{est}$.
Specifically, we utilize the alpha-blending rendering in 3DGS for depth rasterization, where the z-buffer from the ordered Gaussians contributing to a pixel is accumulated for producing the depth value:
\begin{equation}
    d=\sum_{i=1}^nd_i\alpha_i\prod_{j=1}^{i-1}(1-\alpha_j).
\end{equation}
Here $d_i$ represents the z-buffer of the $i$-th Gaussians.
This implementation enables the depth correlation loss.

Finally, we can summarize the training loss:
\begin{equation}
    \mathcal{L}=\lambda_1\mathcal{L}_1(\mathbf{I},\mathbf{\hat{I}})+\lambda_2\mathcal{L}_\text{ssim}(\mathbf{I},\hat{\mathbf{I}})+\lambda_3
    \mathcal{L}_{reg}(\mathbf{D}, \mathbf{D}_{\mathrm{est}}),
\end{equation}
where $\hat{\mathbf{I}}$ is the rendered image and $\mathbf{I}$ is the ground-truth image.
$\mathcal{L}_1$, $\mathcal{L}_{ssim}$ stands for the photometric loss term, and $\mathcal{L}_{reg}$ represents the geometric regularization term on the training views.
$\lambda_1$, $\lambda_2$, and $\lambda_3$ are hyperparameters.

\begin{figure}[t]
    \centering
    \includegraphics[width=0.9\linewidth]{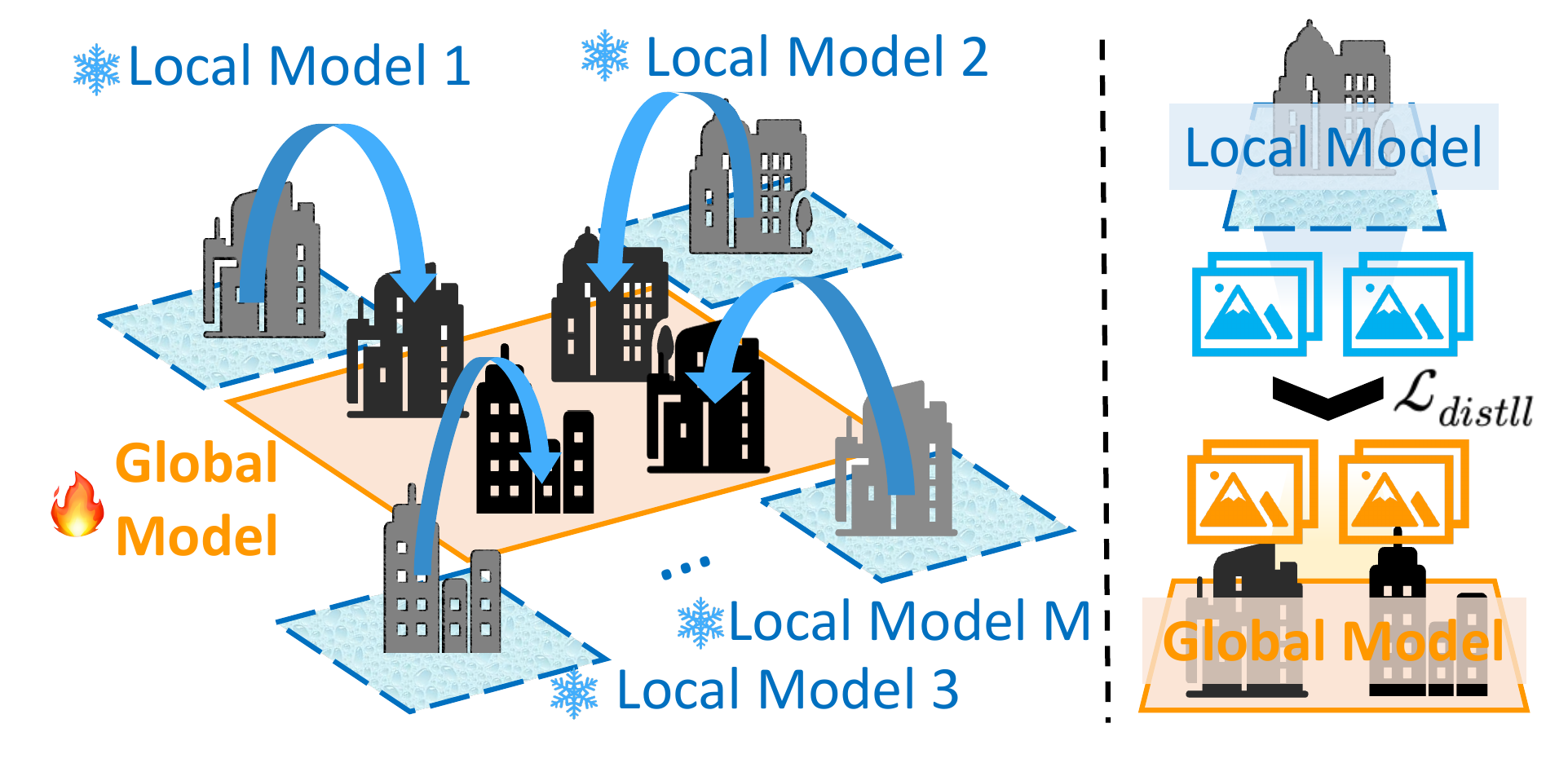}
    \caption{Overview of our Model Aggregation Algorithm.}
    \label{fig:maa}
    \vspace{-3mm}
\end{figure}

\begin{figure}[t]
    \centering
    \includegraphics[width=1\linewidth]{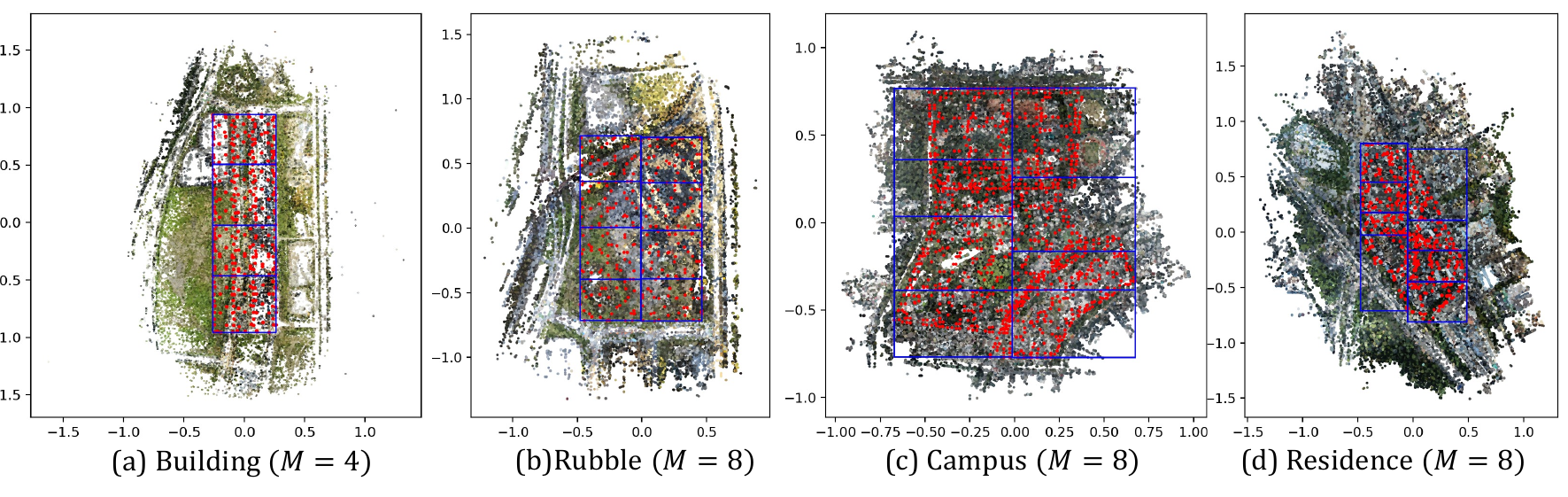}
    \caption{Partitions of different scenes. `\textcolor{blue}{$\Box$}' denote partition areas for each drone. `\textcolor{red}{$\cdot$}' denote camera positions. The background is the sparse points produced by global COLMAP.}
    \label{fig:mask}
    \vspace{-5mm}
\end{figure}

\begin{table*}[htbp]
\centering
\caption{\textbf{Quantitative results of novel view synthesis on Mill19~\cite{turki2022mega} dataset 
      and UrbanScene3D~\cite{lin2022capturing} dataset}. $\uparrow$: higher is better, $\downarrow$: lower is better.
      The \colorbox{red}{red}, \colorbox{orange}{orange} and \colorbox{yellow}{yellow} colors respectively denote the best, the second best, and the third best results on the sparse-view setting.
      The \underline{Underline} denotes the best results in all methods.
    $\dagger$ denotes half of the test images are included in the training set.
    $\ddagger$ denotes it uses all dense images as the training set.}
\resizebox{\linewidth}{!}{
    \begin{tabular}{l|l|ccc|ccc|ccc|ccc}
        \toprule[1.1pt]
         &   &   \multicolumn{3}{c}{\emph{Building}}  
         &   \multicolumn{3}{c}{\emph{Rubble}} 
         &   \multicolumn{3}{c}{\emph{Campus}} 
         &  \multicolumn{3}{c}{\emph{Residence}} \\
         \cmidrule(r){3-5} \cmidrule(r){6-8} \cmidrule(r){9-11} \cmidrule(r){12-14}  
        & & PSNR$\uparrow$ &  SSIM$\uparrow$ & LPIPS$\downarrow$   
        & PSNR$\uparrow$ &  SSIM$\uparrow$ & LPIPS$\downarrow$ 
        & PSNR$\uparrow$ &  SSIM$\uparrow$ & LPIPS$\downarrow$   
        & PSNR$\uparrow$ &  SSIM$\uparrow$ & LPIPS$\downarrow$   \\
        \midrule
        \multirow{3}{*}{\rotatebox[origin=c]{90}{\textbf{Central.}}} 
        & NeRF~\cite{mildenhall2021nerf} $\ddagger$
         & \underline{19.54} & 0.525 & 0.512 
         & 21.14 & 0.522 & 0.546 
         & 19.01 & 0.593 & 0.488 
         & \underline{21.83} & 0.521 & 0.630 \\ 
         \cmidrule(r){2-14}
        & 3DGS~\cite{kerbl20233d}  
         & \cellcolor{orange}16.96 & \cellcolor{red}0.546 & \cellcolor{yellow}0.501 
         & \cellcolor{red}22.57 & \cellcolor{red}\underline{0.637} & 0.449
         & \cellcolor{yellow}14.98 & 0.432 & 0.688 
         & 17.39 & \cellcolor{red}\underline{0.715} & \cellcolor{red}\underline{0.335}  \\

        & DNGaussian~\cite{li2024dngaussian} 
         & \cellcolor{yellow}16.11 & 0.461 & 0.583 
         & 20.72 & 0.550 & 0.540
         & 14.54 & 0.403 & 0.696 
         & \cellcolor{orange}19.11 & 0.656 & 0.393  \\
         
        \midrule
        \multirow{6}{*}{\rotatebox[origin=c]{90}{\textbf{Distributed}}} 
        & Mega-NeRF$\dagger$~\cite{turki2022mega}
        & 19.25 & 0.467 & 0.578 
        & \underline{23.12} & 0.508 & 0.569 
        & \underline{22.58} & 0.507 & 0.669 
        & 21.12 & 0.582 & 0.479 \\

        & Drone-NeRF$\ddagger$~\cite{jia2024drone}
        & 18.46 & 0.490 & 0.469 
        & 19.51 & 0.528 & 0.489 
        & - & - & -
        & - & - & - \\
        
        & Fed-3DGS$\ddagger$~\cite{suzuki2024fed3dgs}
        & 18.66 & \underline{0.602} & 0.362
        & 20.62 & 0.588  & 0.437
        & 20.00 & \underline{0.665} & \underline{0.344}
        & 21.64 & 0.635 & 0.436 \\ 
            \cmidrule(r){2-14}
        & D-3DGS~\cite{kerbl20233d}  
         & 14.49 & 0.434 & 0.517 
         & \cellcolor{yellow}21.70 & \cellcolor{yellow}0.578 & \cellcolor{yellow}0.442 
         & 14.59 & \cellcolor{orange}0.453 & \cellcolor{yellow}0.619
         & \cellcolor{yellow}18.66 & \cellcolor{yellow}0.672 & \cellcolor{yellow}0.347  \\
        
        & VastGaussian~\cite{lin2024vastgaussian}   
        & 15.65 & \cellcolor{yellow}0.467 & \cellcolor{orange}0.420 
        & 19.17 & 0.558 & \cellcolor{orange}0.414 
        & \cellcolor{orange}17.35 & \cellcolor{red}0.558 & \cellcolor{orange}0.540 
        & 17.39 & 0.623 & 0.356 \\

        & DGTR (Ours)     
         & \cellcolor{red}18.47 & \cellcolor{orange}0.532 & \cellcolor{red}\underline{0.392} 
         & \cellcolor{orange}21.72 & \cellcolor{orange}0.591 & \cellcolor{red}\underline{0.360} 
         & \cellcolor{red}19.95 & \cellcolor{yellow}0.517 & \cellcolor{red}0.520 
         & \cellcolor{red}19.97 & \cellcolor{orange}0.677 & \cellcolor{orange}0.339 \\
        \bottomrule[1.1pt]
    \end{tabular}
}
\label{tab:compare}
\vspace{-5mm}
\centering
\end{table*}

\subsection{Model Aggregation}\label{sec:fa}
To obtain the global model from the distributed trained models, we propose a distillation-based model aggregation algorithm that seamlessly stitches all local models together, shown in Fig.~\ref{fig:maa}.
Specifically, we filter out the 3D Gaussians that are outside their original regions and then merge the 3D Gaussians from these non-overlapping regions to obtain a global model, $\mathbb{G}_g$, defined as $\mathbb{G}_g^* = \mathbb{G}_1^* \cup \ldots \cup \mathbb{G}_M^*$, where $\mathbb{G}_m^*$ denotes the filtered model of device $m$.
Next, we collect synthetic views $\mathcal{\overline{I}}_m = \{\mathbf{\overline{I}}^1,\ldots, \mathbf{\overline{I}}^{N_m}\}$ by rendering each device's model (i.e., teacher model) training views from their spares cameras. 
Finally, this set of synthetic views is used to perform a few distillation training epochs on the global model (i.e., student model), using the following distillation loss $\mathcal{L}_{distll}=\lambda_1\mathcal{L}_1(\mathbf{\overline{I}},\mathbf{\hat{I}})+\lambda_2\mathcal{L}_\text{ssim}(\mathbf{\overline{I}},\hat{\mathbf{I}}).$
This method not only quickly aggregates all the device models but also enhances the quality at the boundaries between models.

\section{Experiments}
\subsection{Experimental Settings}
\subsubsection{Datasets}
We simulate the few-shot vast scene reconstruction scenario in mainstream aerial datasets, such as Mill-19~\cite{turki2022mega} and UrbanScene~\cite{lin2022capturing}.  
For Mill-19, we conduct our benchmark on Rubble and Building, each of them contains 1,678 and 1,940 images. 
For UrbanScene, we select Campus and Residence with 5,871 and 2,582 images, respectively.
However, these image sets are too dense for sparse-view training, to this end, we adopt key-frame sampling from~\cite{sun2021neuralrecon} to extract essential images from original scenes, resulting in 297, 297, 1,089, and 756 images for Rubble, Building, Campuse, and Residence, respectively.
As for evaluation, we keep the same settings as previous large-scale scene reconstruction methods like VastGaussian~\cite{lin2024vastgaussian}, including the evaluation views and metrics. 
To simulate the multi-drones multi-drones collaborative reconstruction, we partition the whole scenes into \(M\) regions with no overlap areas. The partition overview is shown in Fig.~\ref{fig:mask}.

\begin{figure}[htbp]
    \centering
    \includegraphics[width=\linewidth]{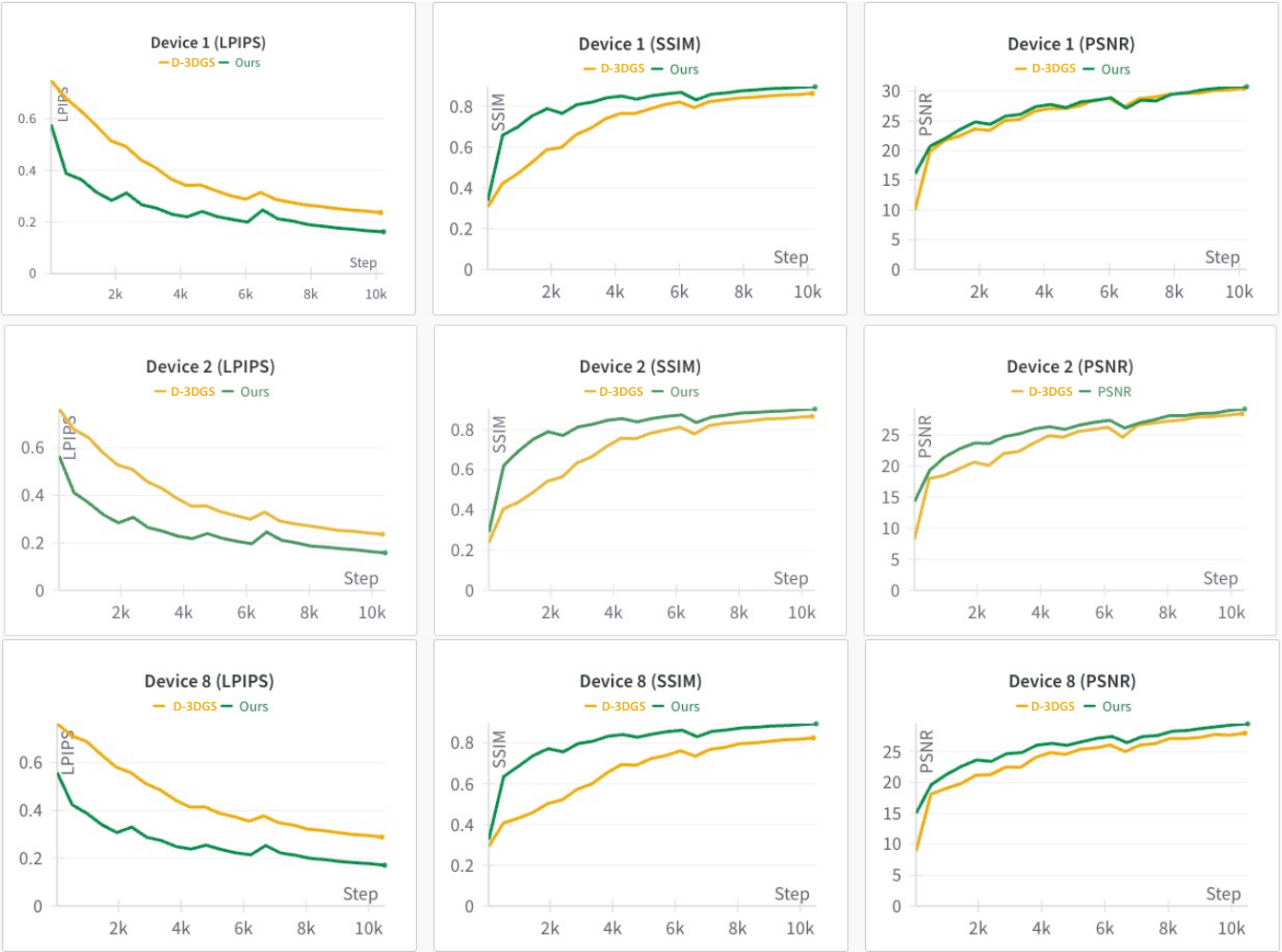}
    \caption{ Training curves for both ours and distributed-3DGS on three devices (i.e., \#1, \#2, \#8). }
    \label{fig:curver}
    \vspace{-3mm}
\end{figure}

\subsubsection{Baselines}
We compare our DGTR with several centralized and distributed methods on four scenes, including NeRF~\cite{mildenhall2021nerf}, 3DGS~\cite{kerbl20233d}, DN-Gaussian~\cite{li2024dngaussian}, Mega-NeRF~\cite{turki2022mega}, Drone-NeRF~\cite{jia2024drone}, Fed-3DGS~\cite{suzuki2024fed3dgs}, modified distributed 3D-GS (D-3DGS), and VastGaussian~\cite{lin2024vastgaussian}. We report the average PSNR, SSIM, and LPIPS scores for all the methods.

\begin{figure*}[htbp]
    \centering
    \includegraphics[width=\linewidth]{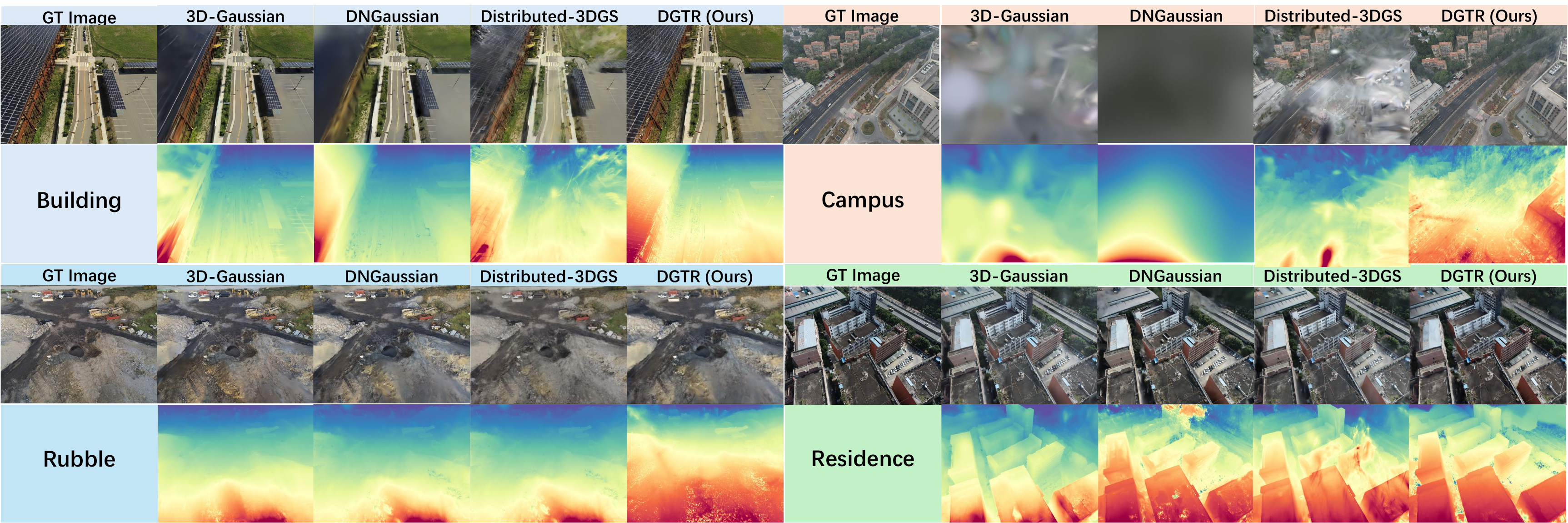}
    \caption{Qualitative results on on Mill19~\cite{turki2022mega} dataset 
          and UrbanScene3D~\cite{lin2022capturing} dataset. }
    \label{fig:vis_overview}
    \vspace{-5mm}
\end{figure*}

\subsubsection{Implementation Details}
Our method is implemented using Pytorch, and all experiments are conducted on A100 40GB GPU. 
For Gaussian initialization, our sliding window step is set to 2, and the optimization step of global alignment is set to 500.
We also set \(\lambda_1=0.8\), \(\lambda_2=0.2\), and \(\lambda_3=0.05\) as the supervision hyper-parameters.
During Device Parallel Training, we train each device for 10,000 steps and densify the Gaussians every 300 iterations to reduce the memory cost. 
During Model Aggregation, we cancel the densification strategy and train the global model for 5 epochs.
For centralized methods, we train them for 50,000 steps to ensure they present the final performance.
For distributed methods, we train each of their devices for 10,000 steps.

\subsection{Comparative Results Analysis}
\subsubsection{Novel View synthesis}
Table~\ref{tab:compare} reports the performance of DGTR and baseline methods in 4 scenes under sparse-view settings. 
A very significant improvement can be found in that our method achieves state-of-the-art in both centralized and distributed methods. 
Moreover, our method achieves comparable performance even compared with some methods trained with dense images.
Additionally, a qualitative comparison can be observed in Fig.~\ref{fig:overview}, our method achieves better visual detail representation (the solar pan on the rooftop in `Building') and geometry accuracy. 
These results demonstrate that our method tends to obtain better generalization to achieve high-quality novel-view synthesis, which benefits from our Gaussian initialization and depth regularisation supervision.

\subsubsection{Convergence}
To demonstrate the advantages of DGTR’s high-quality Gaussian primitives, we present the metric convergence on the evaluation set during training.
Fig.~\ref{fig:curver} illustrates the PSNR, SSIM, and LPIPS convergence on devices \#1, \#2, and \#8 for D-3DGS and DGTR.
In the initial stage, DGTR outperforms D-3DGS across all three metrics, with the most significant improvement seen in LPIPS, where it leads by approximately 20\%.
Additionally, DGTR can achieve better performance than D-3DGS with fewer training steps, enhancing training efficiency.

\subsubsection{Latency}
We compare the latency of DGTR with 3DGS~\cite{kerbl20233d}, which trains the entire scene in a centralized manner, and VastGaussian~\cite{lin2024vastgaussian}, which partitions the entire scene and trains it across multiple GPUs.
Fig.~\ref{fig:latency} illustrates the latency breakdown of each method.
Thanks to the feed-forward model, DGTR has a significant advantage in initialization latency, especially in large-scale scenes (e.g., Campus and Residence). 
Specifically, DGTR can achieve up to $7.42\times$ faster initialization compared to other methods. 
Overall, it is up to $2.06\times$ faster than the other methods.

\begin{figure}[t]
    \centering
    \vspace{-2mm}
    \includegraphics[width=1\linewidth]{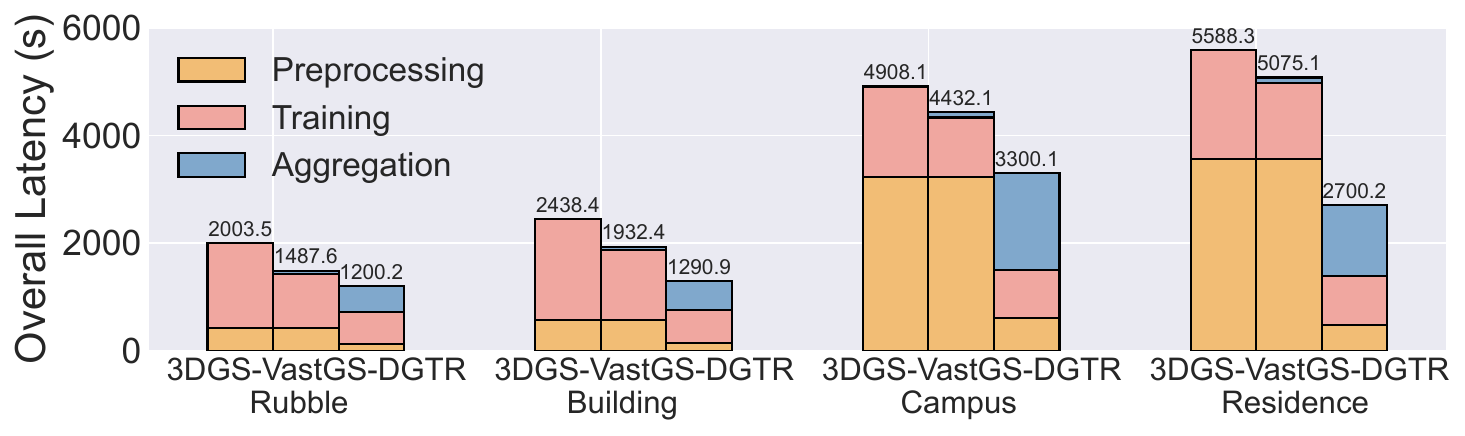}
    \caption{ The overall latency comparison of our method, 3DGS~\cite{kerbl20233d} and VastGaussian (VastGS)~\cite{lin2024vastgaussian}. }
    \label{fig:latency}
    \vspace{-3mm}
\end{figure}

\begin{table}[b]
    \centering
    \caption{Ablations of \(\mathcal{L}_{reg}\).}
    \begin{tabular}{ccc}
    \toprule
         Depth Loss &  Shape Freezing & PSNR\(\uparrow\) \\
    \midrule
    $\times$ & $\times$    &  20.69  \\
    $\checkmark$   & $\times$    &  21.53 \\
    $\checkmark$   & $\checkmark$      &  21.72 \\
    \bottomrule
    \end{tabular}
    \vspace{-5mm}
    \label{tab:ablation_depth}
\end{table}

\begin{table}[b]
    \centering
    \vspace{4pt}
    \caption{Ablations of the scale \(\mathbf{S}\) optimization of \(\mathbf{\Sigma}\).}
    \begin{tabular}{ccc}
    \toprule
         Global Opt. &  Local Opt. & PSNR\(\uparrow\) \\
    \midrule
    $\times$ & $\times$    &  5.19  \\
    $\checkmark$   & $\times$    &  15.06 \\
    $\checkmark$   & $\checkmark$      &  17.53 \\
    \bottomrule
    \end{tabular}
    \vspace{-5mm}
    \label{tab:vis_scale}
\end{table}

\subsection{Ablations}
\subsubsection{Gaussian Initialization}
As shown in Fig. \ref{fig:vis_init}, our initialization process achieves much better visual synthesis results compared with COLMAP by 6+ dB in PSNR. Here we conduct ablations in terms of the scales in covariance attributes, as shown in Fig.~\ref{fig:vis_scale} and Tab.~\ref{tab:vis_scale}. It reveals that global optimization of scale optimizes the scale into poses-level and the local optimization optimizes the relative scale between near and far regions, resulting in massive PSNR improvements by 9.87 dB and 2.47 dB.
\subsubsection{Depth Regularisation} We conduct ablations of our depth regularisation method on the Rubble Scene. As shown in Tab. \ref{tab:ablation_depth}, our depth supervision achieves better results in novel-view synthesis. Additionally, to avoid overfitting the depth maps, we freeze the covariance \(\Sigma\), which shows promising improvements in the final results.

\begin{figure}[t]
    \centering
    \includegraphics[width=\linewidth]{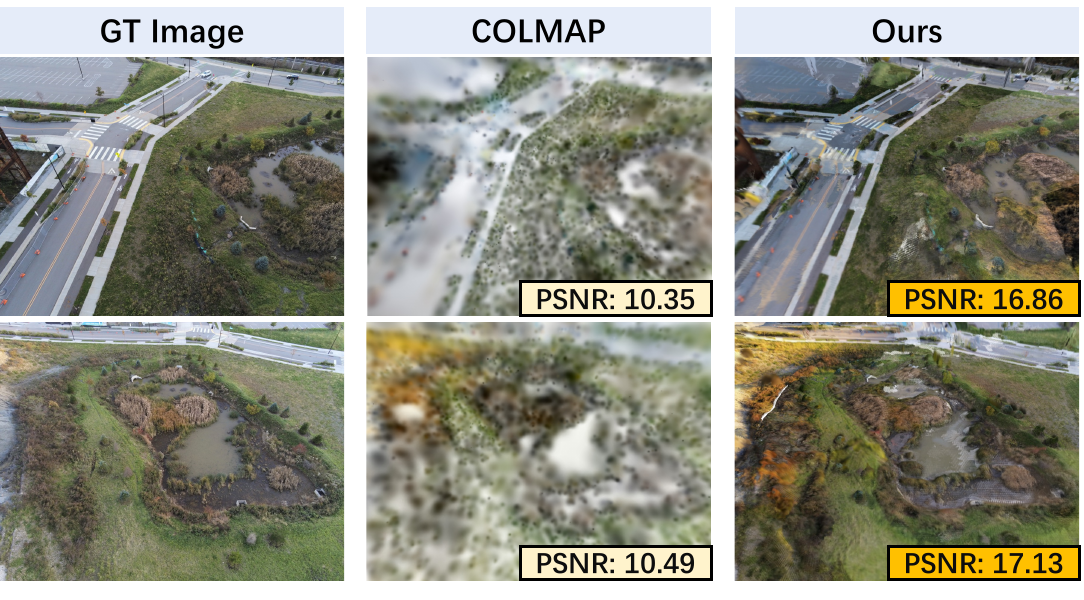}
    \caption{Initialization comparison by COLMAP v.s. Ours.}
    \label{fig:vis_init}
    \vspace{-3mm}
\end{figure}

\begin{figure}[t]
    \centering
    \includegraphics[width=\linewidth]{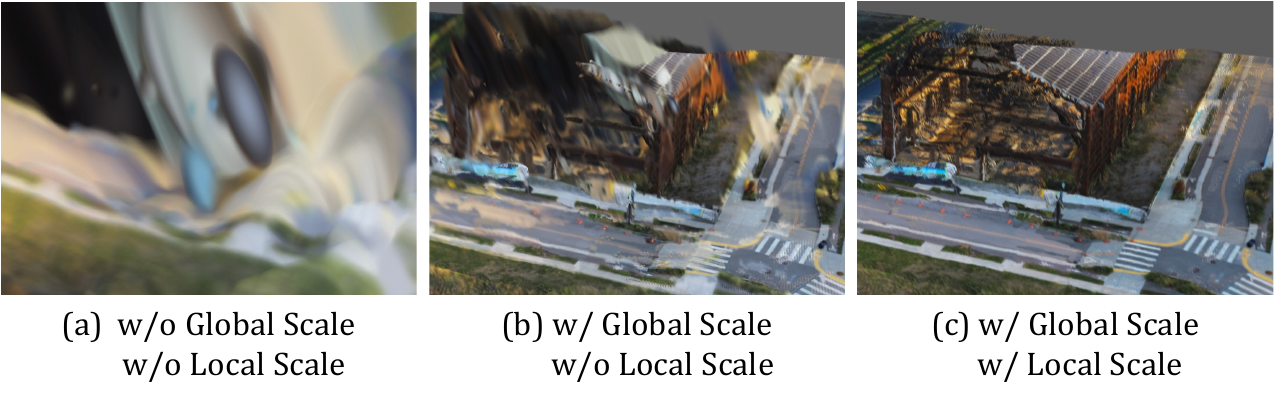}
    \caption{Visual comparisons in scale \(\mathbf{S}\) optimization of \(\mathbf{\Sigma}\). }
    \label{fig:vis_scale}
    \vspace{-3mm}
\end{figure}







\section{Conclusion}
We introduced DGTR, a novel distributed framework for few-shot Gaussian splatting in large-scale scene reconstruction.
Our meticulously designed pipeline comprises three components: distributed feed-forward Gaussian initialization, multiple devices parallel training, and distillation-based aggregation.
With limited training time, our method achieves significant performance improvements compared with other methods.
We also provided a sparse-view vast scene benchmark and demonstrated that DGTR outperforms state-of-the-art methods in both accuracy and overall latency.

\clearpage
\bibliographystyle{IEEEtran}
\bibliography{IEEEfull}

\end{document}